\documentclass[a4paper,fleqn]{cas-dc}

\usepackage[numbers]{natbib}
\usepackage{subfigure}
\def\tsc#1{\csdef{#1}{\textsc{\lowercase{#1}}\xspace}}
\tsc{WGM}
\tsc{QE}
\tsc{EP}
\tsc{PMS}
\tsc{BEC}
\tsc{DE}

\begin{document}
\begin{sloppypar}
\let\printorcid\relax
\let\WriteBookmarks\relax
\def\floatpagepagefraction{1}
\def\textpagefraction{.001}
\pagestyle{empty} 
\shorttitle{Temporal Consistency Two-Stream CNN for Human Motion Prediction}
\shortauthors{Jin Tang et~al.}

\title [mode = title]{Temporal Consistency Two-Stream CNN for Human Motion Prediction}                      

%

\author[]{Jin Tang}[type=editor,
                        auid=000,
                        bioid=1,
]
\cormark[1]
\ead{tangjin@bupt.edu.cn}


\address{School of Artificial Intelligence, Beijing University of Posts and Telecommunications, Beijing, China.}

\author[]{Jin Zhang}
\ead{jinzhang@bupt.edu.cn}

\author[]{Jianqin Yin}[%
   ]
\cormark[1]

\ead{jqyin@bupt.edu.cn}

\credit{Data curation, Writing - Original draft preparation}


%

\cortext[cor1]{Corresponding author}

\nonumnote{This work was supported partly by the Fundamental Research Funds for the Central Universities(Grant No. 2020XD-A04-1).
  }

\begin{abstract}
Fusion is critical for a two-stream network. In this paper, we propose a novel temporal fusion (TF) module to fuse the two-stream joints' information to predict human motion, including a temporal concatenation and a reinforcement trajectory spatial-temporal (TST) block, specifically designed to keep prediction temporal consistency. In particular, the temporal concatenation keeps the temporal consistency of preliminary predictions from two streams. Meanwhile, the TST block improves the spatial-temporal feature coupling. However, the TF module can increase the temporal continuities between the first predicted pose and the given poses and between each predicted pose. The fusion is based on a two-stream network that consists of a dynamic velocity stream (V-Stream) and a static position stream (P-Stream) because we found that the joints' velocity information improves the short-term prediction, while the joints' position information is better at long-term prediction, and they are complementary in motion prediction. Finally, our approach achieves impressive results on three benchmark datasets, including H3.6M, CMU-Mocap, and 3DPW in both short-term and long-term predictions, confirming its effectiveness and efficiency.
\end{abstract}

%


\maketitle

\section{Introduction}

Human motion prediction has been a classic task in the field of computer vision and robotics. It demands predicting the future motion postures of the human body by observing the previous motion sequence, which helps the robots judge the intention of humans more accurately and achieve more excellent human-machine interaction applications \cite{ref1, ref2, ref3}. Unlike motion recognition \cite{ref4, ref5}, which only demands modeling the semantic information of the human body, the challenge of human motion prediction is modeling the dynamics of the human body.

Most of the earliest traditional methods that deal with the human motion prediction task have adopted hidden Markov models \cite{ref6} and linear dynamic systems \cite{ref7}, and so on. They work well in "Walking", "Eating" and other regular movements with high repetition patterns. However, it is challenging to learn the pattern of complex actions such as "Walking Dog", "Posing", and so on. After the emergence of deep learning methods \cite{ref8, ref9}, models can represent high dimensional information, like features from different convolution layer extraction or derivative operated inputs. How to efficiently use and fuse these high-dimensional features has become a significant issue.

For enriching the dynamic temporal information, the optical flow information is used as dynamic temporal information at the pixel level in the video recognition. Sarma et al. \cite{ref10} modeled the optical flow as an additional stream to model the target's motion characteristics in the image. And Wang et al. \cite{ref11} considered the human motion velocity as one of three inputs to feed into the network to model the human body dynamics for motion prediction. Therefore, we believe that the velocity may have advantages in modeling the joint-level motion dynamics.

Two-stream network \cite{ref15, ref16} has become a common network structure to introduce additional information. In this structure, the features are extracted from the two streams separately and then fused together to get output so that the information contained in the fusion features is more prosperous than that in every single stream. Therefore, constructing an effective fusion has become the main problem to be solved in the two-stream network.

For feature fusing, many works \cite{ref12, ref13, ref14, ref15, ref16} propose to apply multi-scale features or high-dimensional information to model human dynamics. Some \cite{ref12, ref13, ref15, ref16} use feature addition to add features from a residual connection. Besides, Li et al. \cite{ref14} used convolution layers to fuse the concatenation output of multi-scale features. However, as shown in Fig. 1(a)(b), methods like concatenation or addition neglect spatial-temporal co-occurrence and continuity of time. In this paper, we design a novel temporal fusion module that can ensure the coupling of the two-stream information by restoring the two channels of features' time dimension, as shown in Fig. 1(c). 

\begin{figure*}[htbp]
    \centering
    \subfigure{
        \includegraphics[width=1\textwidth]{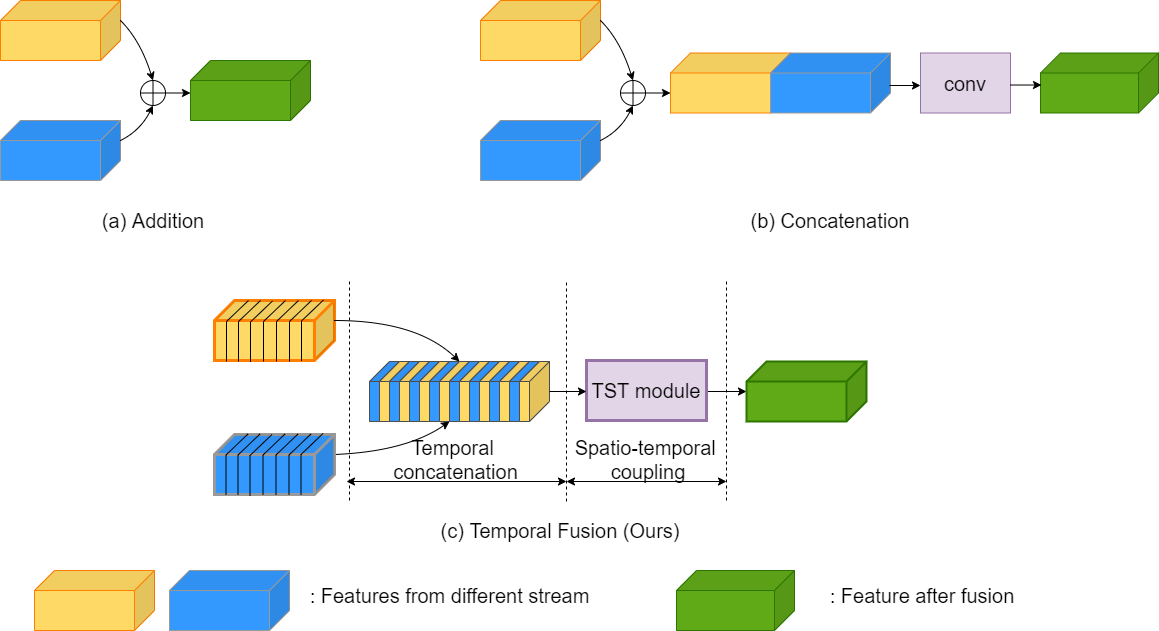}
    }
\end{figure*}
\begin{table*}
\begin{flushleft}
\textbf{Fig.1:} Methods for feature fusion. (a) Addition of two-stream features. (b) Concatenation of two-stream features. (c) A temporal fusion module based on temporal concatenation and a reinforcement trajectory spatial-temporal (TST) module. When inputting features from different modeling streams, (a)(b) neglect the temporal consistency comparing with (c).
\end{flushleft}
\end{table*}

As noted above, we model the velocity and position prediction from two different streams respectively. And the future velocity sequence of human joints generated by the V-stream will be restored into the position space to maintain its physical significance equal to that of position space. Then, we conduct the TF module for fusing position space output and velocity space output so as to take advantage of the dynamic information of human motion and give the ultimate prediction results. The temporal concatenation fuses V-stream features with P-stream features separately by the dimension of time, which achieves better integration of spatiotemporal information comparing with feature addition. In this way, each time-step's P-stream features and V-stream features are fused separately to ensure temporal consistency. To solve the problem of poor coupling after fusion, we use a TST module to continue modeling the spatiotemporal coupling features and generate the ultimate prediction results.

In summary, the main contributions are two-fold: (1) The novel temporal fusion (TF) module that keeps the temporal consistency and enhances the spatial-temporal features coupling. (2) A two-stream framework that integrates the advantages of velocity stream in short-term modeling and position stream in long-term modeling. Our approach is also general and easy to incorporate into a two-stream-based prediction framework. Our experiments on three standard human motion prediction benchmarks evidence the benefits of our approach.

The remainder of the paper is organized as follows. The next section investigates the related work. Section 3 discusses our model in detail. The dataset, evaluation criteria, the experimental-based comparisons of different methods, and ablation studies are presented in Section 4. Finally, conclusions and future work directions are stated in section 5. Our code will be made public after the paper is accepted.

\section{Related work}

Since human motion can be regarded as a temporal sequence, recurrent neural network (RNN) methods \cite{ref12, ref17, ref18, ref19, ref20} have achieved quite a remarkable effect on human motion prediction. Gopalakrishnan et al. \cite{ref17} proposed a two-stage processing RNN architecture to capture the difference in the power spectrum of the ground truth frames, which helped reduce an accumulation of next-step error generated recursively. Along with Fragkiadaki et al. \cite{ref19} proposed to use Encoder-Recurrent-Decoder (ERD) to extend previous long-short-term memory (LSTM) models in the literature to jointly learn representations and their dynamics. However, \cite{ref17, ref18, ref19} neglect the problem of discontinuities between the observed poses and the predicted future ones. Martinez et al. \cite{ref13} used a sequence-to-sequence residual RNN to solve discontinuities. However, it still neglected the co-occurrence of the spatial-temporal relationship from a human motion sequence. The recursive modeling ability of the RNN network is excellent, but most of the works that performing RNN ignore the correlation between the spatial information of human joints and time. Liu et al. \cite{ref20} proposed hierarchical motion context modeling to update the local skeletal state and introduced the spatial connection between joint points into an RNN. However, the proposed network loses sight of the global temporal co-occurrence relationship of the motion sequence.

Instead of RNN, convolutional neural networks (CNN) achieve better performance in many works \cite{ref14, ref21, ref22}. They adopt a sequence to sequence model that performs convolution operation in the temporal domain so that the global temporal feature of human motion correlation can be captured. Li et al \cite{ref14} used a long-term convolutional encoder to capture the long-term information. Spatial-temporal features are fed into the next layer at a different level to model the motion dynamic effectively. Liu et al. \cite{ref21} also proposed a high-efficiency spatial-temporal modeling network, noted as TrajectoryNet. TrajectoryNet is a feed-forward network, treating the human motion as a composition of joint-level trajectories and modeling them by designing joints in a natural skeletal order, which retained both temporal and spatial features. Both works in \cite{ref14, ref21} model human dynamics in position space, conducting 3D joint-level coordinates as input and leaving high dimensional space input to be discussed.

Optical flow \cite{ref23, ref24, ref25, ref26} is widely used high-dimensional information for the task of video action recognition. It searches the corresponding relationship between the previous frame and the current frame by the difference of pixels between adjacent frames, which can be denoted as the pixel-wise velocity. As for joint-wise, recently, many works \cite{ref11, ref12, ref13, ref18, ref19, ref20, ref27, ref28} choose to model human motion dynamics by multi-level dynamic information to predict the regular movement changes of human motion and achieved well-performance prediction. Gui et al. \cite{ref18} proposed an adversarial geometry-aware encoder-decoder (AGED) to perform adversarial training at the sequence level, which adopts the idea of multi-level GAN. The discriminators designed in AGED are used to discriminate the fidelity and continuity of predicted sequence separately. And Gui et al. \cite{ref27} proposed to adopt proactive and adaptive meta-learning (PAML) to jointly learns a generic model initialization and an effective model adaptation strategy. \cite{ref11} directly points out that the velocity of the input sequence can be presented as input to better model human dynamism by an RNN network. Human pose, pose velocity and position embedding are fed into RNN composed by GRU to encode different frames' absolute temporal positions and preserve motion continuities. Shu et al. \cite{ref28} proposed a spatiotemporal co-attention RNN to model dynamics in skeleton motion and joint motion corresponding to temporal evolution and spatial coherence. In summary, multi-level dynamic information can be explored to enrich learning features for different tasks.

\begin{figure*}[htbp]
    \centering
    \subfigure{
        \includegraphics[width=0.9\textwidth]{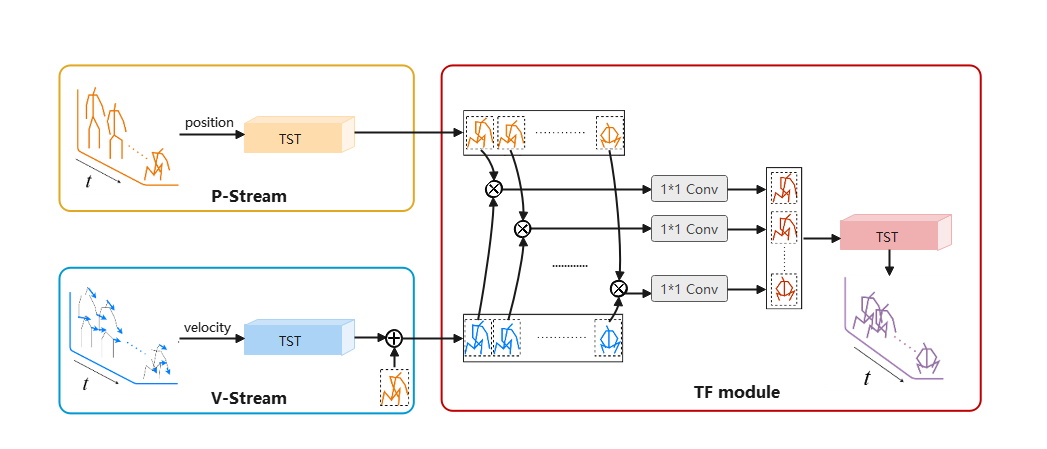}
    }
    \label{fig:fig_micPerMon}
\end{figure*}
\begin{table*}
\begin{flushleft}
\textbf{Fig.2:} Overall Architecture. The model of velocity is also important for skeleton-based motion prediction but is neglected in most earlier works. In this work, a V-Stream and a P-Stream are used to model velocity and position respectively. The TF module is composed of temporal concatenation and a reinforcement TST block.
\end{flushleft}
\end{table*}

In many works \cite{ref10, ref11, ref15, ref16}, two-stream or multi-stream modeling networks are proposed to effectively integrate high-dimensional or multi-level information into position space features and achieve better recognition or prediction results. Shi et al. \cite{ref16} proposed a two-stream GCN-based framework to model both the bone and joint information respectively to increase the flexibility of the model for graph construction and bring more generality to adapt to various data samples. As well as Wang et al. \cite{ref11} designed a Position-Velocity RNN (PVRNN) that took in three inputs and predicts pose velocities, which are then added to the previous posts to get the future poses. The spatial-temporal co-occurrence of the learned feature is obviously ignored. Then the predictions of pose velocities and human poses are used as input for the next time step in the RNN modeling process. Furthermore, \cite{ref10, ref15, ref16} also lose sight of the part of fusion. The human motion sequence features are simply fused by linear options. Inspired by this, this paper, based on CNN for spatial-temporal modeling, proposes a fusion module suitable for human motion temporal sequence, to fuse the features from position space and velocity space generated by a two-stream network.  

\section{Methodology}

The human motion prediction task is to observe the input pose sequence, which is denoted as $S=\{P_1,P_2,…,P_{t_i}\}$ along the dimension of time, and generate future pose sequence generated which is denoted as $S'=\{P_{{t_i}+1}, P_{{t_i}+2}, …, P_{{t_i}+{t_o}}\}$. The $i$th pose $P_i$ is composed of joints, which can be denoted as $P_i=\{J_1,J_2,…,J_n\}$. Index $n$ means the number of human skeletal joints. Meanwhile, the $k$th joint can be represented as $J_k=\{x_k,y_k,z_k\}$ in 3D coordinate space.

However, for the specific task of human motion predicting, there are a few common pitfalls that we would like to improve like the predicted pose tend to converge to the mean pose or fail to generate natural-looking poses due to clear discontinuities in the first frame. In this paper, we attack the above problems in a variety of ways. First, we introduce the velocity feature vector, computed by a time-step position difference, feeds into predictive neural models, which naturally holds local temporal information that is crucial when generating smooth and consistent motion trajectories. We predict the human pose by adding the predicted velocity on the last observed frame. We observe that naive input replacement cannot achieve high performance in long-term prediction even though it effectively improves the short-term prediction and gets more natural-looking poses. Furthermore, we then propose a two-stream convolutional network that consisting of a dynamic velocity stream and a static position stream, as shown in Fig. 2. To keep the spatial-temporal coherence, we propose a TF module to fuse prediction pose from P-Stream and V-Stream by the dimension of time, which achieves higher predictive performance in both short-term and long-term predictions. The details of the architecture are discussed below.

\subsection{Architecture}

We use a convolutional network to build the prediction pipeline since convolutional network can better model the spatial-temporal coupling features. The framework of our network consists of two parts, as shown in Fig. 2.

(1) For the first part, we model position space information and velocity space information in two streams using the TST block, respectively. Then each stream gives preliminary predictions. The predicted velocity vector was accumulated to the last frame of position space input to recover the velocity output to position space for fusion.

(2) For the second part, we fuse the prediction results from two streams in chronological order by our temporal concatenation. Since simple proportional fusion by 1*1 convolution still leads to poor spatial-temporal coupling, we use another TST block to continue modeling the spatial-temporal co-occurrence features.

In summary, the TST block is a vital block in our network. We use it in the two-stream modeling phase and TF module. The illustration of the TST is shown in Fig. 3.

\paragraph{Trajectory Spatial-temporal(TST).}The TST block aims to model the trajectory of each input tensor. Since human motion can be regarded as composed of each joint's trajectory, the convolution layer can model its trajectory information by the time dimension as the channel. Firstly, we process H3.6M as 22 key joints in the human body for training which is similar to the settings as the baseline TrajectoryNet \cite{ref21}. Then, According to VGG \cite{ref29}, two 3*3 convolutions have the same receptive field as one 5*5 convolution when stride and padding are set to 1. We employ eleven 3*3 convolution layers in the TST block. The final equivalent receptive field of temporal dimension is enlarged, capable of covering the size of a skeletal pose. Meanwhile, we use six 1*1 convolutional layers to construct residual connections, as shown in Fig. 3. The deeper layer is connected with the lower layer by a residual connection so that the coarse-grained features can be supplied with fine-grained features.

\subsection{Dynamic Velocity Stream (V-Stream).}

Many works have proposed methods of using velocity to model the dynamics of objects. As mentioned in Introduction, optical flow describes the displacement of pixels, which can be regarded as pixel-level velocity. Karen et al. \cite{ref23} used a two-stream network to process video RGB frames and optical flow respectively to give the classification confidence. Since video is a temporal composition of RGB frames, human motion can also be treated as a temporal composition of 3D joint coordinates. Inspired by this, the concept of velocity is applied to represent the joint-level velocity information in our work.

Since the observed data is the joints' position, the motion sequence's velocity should be extracted from the observed position sequence. We adopt the difference between adjacent time-steps of a joint's position to indicate the joint's velocity at the moment. That is to say, the velocity of the kth joint at the moment $t$ $V_{(k,t)}$ can be calculated by Eq [1]:
\begin{equation}
V_{(k,t)}=\{x_{k,t+1}-x_{k,t}, y_{k,t+1}-y_{k,t}, z_{k,t+1}-z_{k,t}\}
\end{equation}
In this case, $V_{k,t}$ captures more dynamic temporal information and discards redundant information from position space at the moment of $t$. The input sequence of velocity can be described as $S_v \{V_1,V_2,…,V_{t_i-1} \}$. As shown in Fig.2, we denote the velocity sequence's shape as $\left( N\times T-1\times 3\right)$ to represent the velocity of each joint of the human body at different moments, and the position sequence's shape as $\left(N\times T\times  3\right)$. $N$ is the number of joints and $T$ is the number of input frames, "$3$" contains 3D coordinates of $x,y,z$. This constructed high-dimensional information will be piped into the V-stream network to generate a prediction that pays more attention to dynamic information. The prediction of velocity can be represented as $\widehat{S_v}=\{\widehat{U_v^1},\widehat{U_v^2},…,\widehat{U_v^{t_o}}\}$, and the predicted velocity of the joints in the $i$th frame $\widehat{U_v^i}$is composed by $\{\widehat{V_{(1,i)}},\widehat{V_{(2,i)}},…,\widehat{V_{(n,i)}}\}$, and the $j$th joint's velocity at the $t$th frame can be represented as $\widehat{V_{(j,t)}}=\{\widehat{\Delta x},\widehat{\Delta y},\widehat{\Delta z}\}$. These deviation vectors are superposed to the last frame of the position sequence to recover it into position space for further fusion. That is, we calculate the ith predicted pose of the velocity stream by Eq [2].
\begin{equation}
\widehat{P_v^i} =\{J_{1,t_i}+\sum_{k=1}^i \widehat{V_{1,t}},J_{2,t_i}+\sum_{k=1}^i \widehat{V_{2,t}},…,J_{n,t_i}+\sum_{k=1}^i \widehat{V_{n,t}}\}
\end{equation}
This preliminary prediction from the velocity stream will be used for feature fusing in the next section.

\begin{figure*}[htbp]
    \centering
    \subfigure{
        \includegraphics[width=0.9\textwidth]{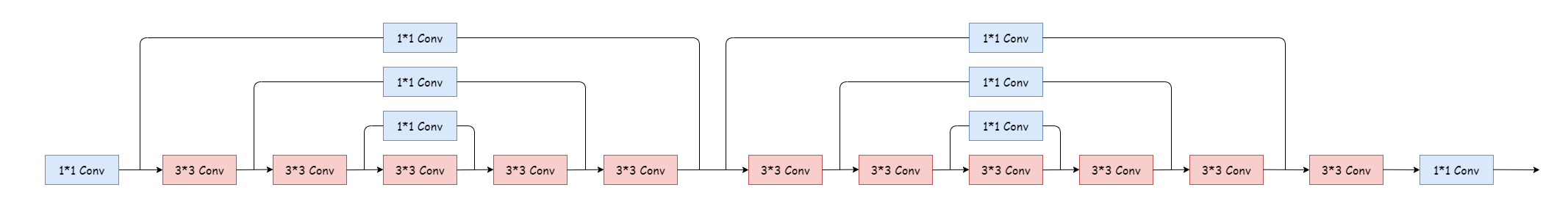}
    }
    \label{fig:fig_micPerMon}
\end{figure*}
\begin{table*}
\begin{flushleft}
\textbf{Fig.3: } Trajectory spatial-temporal (TST) block. It improves the spatial-temporal feature coupling by using residual connections. Each convolution layer is coupling with leaky-Relu and Dropout option.
\end{flushleft}
\end{table*}

\subsection{Temporal Fusion module (TF module)}

Since simple concatenation ignores the temporal correlation between the position space and the velocity space, the TF module comprises a temporal concatenation and a reinforcement trajectory spatial-temporal (TST) network. For temporal concatenation, a typical approach is applied, which is recovering the predicted sequence length $\widehat T$ from the number of channels, then they are fused in chronological order. As shown in Fig. 2, each stream outputs a tensor at the shape of $\left(N\times \widehat T\times 3\right)$ as its prediction before the fusion. Since the tensor could be restored to the time dimension, the fusion based on each moment can be conducted respectively. The prediction results generated from P-stream and V-stream will be respectively divided into $\widehat T$ tensors in chronological order, and then the same time-step tensors are concatenated. In this way, we achieve $\widehat T$ tensors at the size of $\left(N\times 2\times 3\right)$. For each moment's temporal concatenation result, one 1*1 convolution layer as a dynamic selector to balance the prediction from two streams. That is to say, these $\widehat T$ 1*1 convolution layers give a preliminary prediction fusing result of position space and velocity space for each predicting moment. In this way, each selector keeps its weight for leveraging two-stream features. Comparing with many works discussed in \cite{ref10, ref11, ref15, ref16} that fuse two-stream features by concatenation or addition, our method maintains the temporal correlation between the position space and the velocity space.

As shown in Fig.2, the temporal concatenation result is sent to a TST block in the TF module. Since the 1*1 convolution layer only gives the proportional fusion result, there is a lack of spatial coupling that each joint uses the same proportion to calculate the new joint position. In case of this, an additional TST block is added to refine the spatial-temporal coupling and output the final fusion result.

\section{Experiments}

We evaluate our model and compare it with the state-of-the-art on three benchmark datasets, including Human3.6M(H3.6M) \cite{ref30}, the CMU mocap dataset (CMU-Mocap), and the 3D pose in the Wild dataset (3DPW) \cite{ref31}. In the following text, the experiment details, datasets, the evaluation metrics, and the baseline are first introduced, then the results of our method are presented and analyzed, and the predictive performance of our model is also visualized in the last.

\subsection{Datasets}

\paragraph{H3.6M.} H3.6M \cite{ref30} dataset is a commonly used dataset for human motion prediction, captured and recorded by 7 professional actors for training and evaluation. 15 activities are provided by pose sequences. Each frame is represented as 32 joints in relative 3D joint positions. Following existing works \cite{ref14, ref21}, we down-sample the sequences by 2 to 25 frames per second and use Subject 5 as the test data and Subjects 1, 6, 7, 8, 9 as training. Preprocessing and data selection criteria are directly followed from the recent work of \cite{ref14}. 
\paragraph{CMU-Mocap.} CMU-Mocap\footnote{http://mocap.cs.cmu.edu/} provides 2,235 human motion sequences. The frames are recorded by VGA cameras. Following baselines in \cite{ref12, ref21}, eight actions are selected from category "locomotion", "physical activities \& sports", "common behaviors and expressions" and "communication gestures and signals", we use the same dataset splits for training and testing. 
\paragraph{3DPW.} 3DPW \cite{ref31} consists of indoor and outdoor actions such as shopping, doing sports, and hugging, including 60 sequences and more than 51k frames. For a fair comparison, we use the official split sets for experiments.

\begin{table*}
\begin{flushleft}
\textbf{Table 1:} Short-term prediction of 3D coordinates on H3.6M for all actions. Our method outperforms the baselines on average prediction on all time steps.
\end{flushleft}
\end{table*}
\begin{table*}[]\normalsize
\begin{center}
\renewcommand\arraystretch{1}
\renewcommand\tabcolsep{3pt}
\begin{tabular}{l|cccc|cccc|cccc|cccc}
\hline            & \multicolumn{4}{c|}{Walking}                                   & \multicolumn{4}{c|}{Eating}                                     & \multicolumn{4}{c|}{Smoking}                                   & \multicolumn{4}{c}{Discussion}                               \\
Milliseconds              & 80            & 160           & 320           & 400           & 80            & 160           & 320           & 400            & 80            & 160           & 320           & 400           & 80           & 160           & 320           & 400           \\
\hline ConvSeq2Seq\cite{ref14}   & 17.1          & 31.2          & 53.8          & 61.5          & 13.7          & 25.9          & 52.5          & 63.3           & 11.1          & 21.0            & 33.4          & 38.3          & 18.9         & 39.3          & 67.7          & 75.7          \\
LearnTrajDep\cite{ref12}  & 8.9           & 15.7          & 29.2          & \textbf{33.4} & 8.8           & 18.9          & 39.4          & 47.2           & 7.8           & 14.9          & 25.3          & 28.7          & 9.8          & 22.1          & \textbf{39.6} & \textbf{44.1} \\
TrajectoryNet\cite{ref21}          & 8.2           & 14.9          & 30.0            & 35.4          & 8.5           & 18.4          & 37.0            & 44.8           & 6.3           & 12.8          & 23.7          & 27.8          & 7.5          & 20.0            & 41.3          & 47.8          \\
\hline Ours          & \textbf{7.8}  & \textbf{14.8} & \textbf{28.9} & 34.4          & \textbf{7.7}  & \textbf{17.1} & \textbf{34.5} & \textbf{42.6}  & \textbf{6.2}  & \textbf{12.7} & \textbf{23.3} & \textbf{27.6} & \textbf{7.1} & \textbf{18.7} & 39.7          & 46.8          \\
\hline               & \multicolumn{4}{c|}{Directions}                                & \multicolumn{4}{c|}{Greeting}                                   & \multicolumn{4}{c|}{Phoning}                                   & \multicolumn{4}{c}{Posing}                                   \\
Milliseconds               & 80            & 160           & 320           & 400           & 80            & 160           & 320           & 400            & 80            & 160           & 320           & 400           & 80           & 160           & 320           & 400           \\
\hline ConvSeq2Seq\cite{ref14}   & 22.0            & 37.2          & 59.6          & 73.4          & 24.5          & 46.2          & 90.0            & 103.1          & 17.2          & 29.7          & 53.4          & 61.3          & 16.1         & 35.6          & 86.2          & 105.6         \\
LearnTrajDep\cite{ref12}  & 12.6          & 24.4          & \textbf{48.2} & \textbf{58.4} & 14.5          & 30.5          & 74.2          & 89.0             & 11.5          & 20.2          & 37.9          & 43.2          & 9.4          & 23.9          & 66.2          & 82.9          \\
TrajectoryNet\cite{ref21}          & 9.7           & \textbf{22.3} & 50.2          & 61.7          & \textbf{12.6} & 28.1          & 67.3          & 80.1           & 10.7          & 18.8          & \textbf{37.0}   & \textbf{43.1} & 6.9          & \textbf{21.3} & \textbf{62.9} & \textbf{78.8} \\
\hline Ours          & \textbf{9.4}  & 22.9          & 53.7          & 65.1          & 12.7          & \textbf{28.0}   & \textbf{64.6} & \textbf{79.1}  & \textbf{10.0}   & \textbf{18.6} & 37.9          & 44.4          & \textbf{6.8} & 21.6          & 63.5          & 79.9          \\
\hline               & \multicolumn{4}{c|}{Purchases}                                 & \multicolumn{4}{c|}{Sitting}                                    & \multicolumn{4}{c|}{Sitting Down}                              & \multicolumn{4}{c}{Taking Photo}                             \\
Milliseconds               & 80            & 160           & 320           & 400           & 80            & 160           & 320           & 400            & 80            & 160           & 320           & 400           & 80           & 160           & 320           & 400           \\
\hline ConvSeq2Seq\cite{ref14}   & 29.4          & 54.9          & 82.2          & 93.0            & 19.8          & 42.4          & 77.0            & 88.4           & 17.1          & 34.9          & 66.3          & 77.0            & 14.0           & 27.2          & 53.8          & 66.2          \\
LearnTrajDep\cite{ref12}  & 19.6          & 38.5          & 64.4          & 72.2          & 10.7          & 24.6          & 50.6          & \textbf{62.0}    & 11.4          & 27.6          & 56.4          & 67.6          & 6.8          & 15.2          & 38.2          & 49.6          \\
TrajectoryNet\cite{ref21} & \textbf{17.1} & \textbf{36.1} & 64.3          & 75.1          & 9.0             & 22.0            & \textbf{49.4} & 62.6           & \textbf{10.7} & 28.8          & 55.1          & 62.9          & \textbf{5.4} & \textbf{13.4} & \textbf{36.2} & \textbf{47.0}   \\
\hline Ours          & 17.4          & \textbf{36.1} & \textbf{59.4} & \textbf{67.4} & \textbf{8.6}  & \textbf{21.5} & 50.3          & 63.6           & 11.0            & \textbf{27.4} & \textbf{51.2} & \textbf{60.7} & 5.5          & 13.5          & 37.1          & 49.0            \\
\hline               & \multicolumn{4}{c|}{Waiting}                                   & \multicolumn{4}{c|}{Walking Dog}                                & \multicolumn{4}{c|}{Walking Together}                          & \multicolumn{4}{c}{Average}                                  \\
Milliseconds              & 80            & 160           & 320           & 400           & 80            & 160           & 320           & 400            & 80            & 160           & 320           & 400           & 80           & 160           & 320           & 400           \\
\hline ConvSeq2Seq\cite{ref14}   & 17.9          & 36.5          & 74.9          & 90.7          & 40.6          & 74.7          & 116.6         & 138.7          & 15.0            & 29.9          & 54.3          & 65.8          & 19.6         & 37.8          & 68.1          & 80.2          \\
LearnTrajDep\cite{ref12}  & 9.5           & 22.0            & 57.5          & 73.9          & 32.2          & 58.0            & 102.2         & 122.7          & 8.9           & 18.4          & 35.3          & 44.3          & 12.1         & 25.0            & 51.0            & 61.3          \\
TrajectoryNet\cite{ref21} & \textbf{8.2}  & 21.0            & 53.4          & 68.9          & 23.6          & 52.0            & 98.1          & 116.9          & 8.5           & 18.5          & 33.9          & 43.4          & 10.2         & 23.2          & 49.3          & 59.7          \\
\hline Ours          & 8.3           & \textbf{20.5} & \textbf{51.4} & \textbf{66.3} & \textbf{20.8} & \textbf{47.7} & \textbf{94.1} & \textbf{108.7} & \textbf{7.7}  & \textbf{18.1} & \textbf{31.9} & \textbf{41.1} & \textbf{9.8} & \textbf{22.6} & \textbf{48.1} & \textbf{58.4}\\
\hline
\end{tabular}
\end{center}
\end{table*}

\subsection{Implementation Details}

Following the setting and processing in \cite{ref21}, all experiments are carried out in 3D coordinate space. In experiments, all models are implemented by TensorFlow. To be consistent with the literature, we report our results for short-term (< 400ms) and long-term (>400ms) predictions. We follow the same experiment setting in \cite{ref12, ref14, ref21}, giving 10 frames (400 milliseconds) as input to predict 10 or 25 frames for short-term or long-term evaluation on H3.6M and CMU-Mocap. For 3DPW, we use 30 frames (1 second) as long-term prediction. Meanwhile, we use the same skeletal representation, dropout option and activation function as TrajectoryNet \cite{ref21} for modeling local spatial features. All models are trained with Adam optimizer, and the learning rate is initialized to 0.0001. MPJPE (Mean Per Joints Position Error) \cite{ref30} is used as our loss function to train the network, see Eq [3].
\begin{equation}
l=\frac{1}{T_o*N} \sum_{t=1}^{T_o}\sum_{k=1}^N\Vert\widehat{J_{t,k}}-J_{t,k} \Vert^2 
\end{equation}
$J_{t,k}$ denotes the $k$th joint’s 3D coordinates at the $t$th moment from the ground truth. Similarly, $\widehat{J_{t,k}}$ denotes the $k$th joint’s 3D coordinates at the $t$th moment from prediction results. $T_o$ and $N$ represents the length of the output sequence and the number of joints.
MPJPE can measure the average value of Euclidean space errors between the ground truth and the predicted joints and is often used to evaluate human motion estimation and prediction results. We do not adopt MPJPE on the preliminary prediction results generated by the two streams independently because we intend to avoid an excessive degree of feature convergence learned by the two streams. We notice that the regularity of 3DPW outdoor actions is poor, then we set the weight of the last 22 predicted frames to 0.2 for reducing error accumulating while training.

We also make use of the Mean Per Joint Position Error(MPJPE) \cite{ref30} in millimeters as the error evaluation metric on all datasets. As mentioned in \cite{ref14}, angles are not a good representation to evaluate motion prediction. We employ the measurement of the Euclidean distance between the ground-truth pose and our predicted pose in the 3D coordinate space as the error metric. For example, the H3.6M dataset contains several human motion pose sequences. After dividing the training set and the test set, we take $T_i+T_o$ frames of the sequence as a sample. The first $T_i$ frames are input to our two-stream network to obtain the predicted sequence $\widehat S$ which is composed by predicted joint $\widehat J_(t,k)$, and the last $T_o$ frames of the sample consist the ground truth sequence $S$. Then the MPJPE between the $\widehat S$ and the ground truth sequence $S$ is calculated according to Eq[3].

\subsection{Baselines}

We compare our method with three convolutional networks, LearnTrajDep \cite{ref12}, convSeq2Seq \cite{ref14} and TrajectoryNet \cite{ref21}. The MPJPE results are taken from their respective papers. Specifically, LearnTrajDep \cite{ref12} uses a graph convolutional network to learn graph connectivity automatically. ConvSeq2Seq \cite{ref14} proposes a convolutional encoder-decoder network to capture both invariant and dynamic information of human motion. The proposed encoder extracts multi-level spatial-temporal features for forwarding modeling. TrajectoryNet \cite{ref21} proposes to learn human dynamics in trajectory space by CNN modeling and residual connection. We implement the same data preprocessing with the baselines, which takes 3D coordinates as input and output. 

\begin{table*}
\begin{flushleft}
\textbf{Table 2:} Long-term prediction of 3D coordinates on H3.6M for all actions. Our method outperforms the baselines on average prediction on all time steps.
\end{flushleft}
\end{table*}
\begin{table*}[]\small
\begin{center}
\renewcommand\arraystretch{1}
\renewcommand\tabcolsep{0.8pt}
\begin{tabular}{l|cc|cc|cc|cc|cc|cc|cc|cc}
\hline
                      & \multicolumn{2}{c|}{Walking}          & \multicolumn{2}{c|}{Eating}    & \multicolumn{2}{c|}{Smoking}      & \multicolumn{2}{c|}{Discussion}   & \multicolumn{2}{c|}{Directions} & \multicolumn{2}{c|}{Greeting}    & \multicolumn{2}{c|}{Phoning}          & \multicolumn{2}{c}{Posing}      \\
Milliseconds          & 560                  & 1000           & 560           & 1000           & 560             & 1000            & 560             & 1000            & 560            & 1000           & 560             & 1000           & 560               & 1000              & 560            & 1000           \\ \hline
ConvSeq2Seq\cite{ref14}   & 59.2                 & 71.3           & 66.5          & 85.4           & 42.0            & 67.9            & 84.1            & 116.9           &                &                &                 &                &                   &                   &                &                \\
LearnTrajDep\cite{ref12}  & 42.2                 & 51.6           & 57.1          & 69.5           & 32.5            & 60.7            & \textbf{70.5}   & \textbf{99.6}   & 79.6           & 102.9          & 95.8            & 89.9           & 62.6              & 113.8             & \textbf{107.2} & 211.8          \\
TrajectoryNet\cite{ref21} & 37.9                 & \textbf{46.4}  & 59.2          & \textbf{71.5}  & 32.7            & 58.7            & 75.4            & 103             & 84.7           & 104.2          & 91.4            & \textbf{84.3}  & 62.3              & 113.5             & 111.6          & \textbf{210.9} \\
Ours                  & \textbf{37.7}        & 48.3           & \textbf{56.3} & 71.7           & \textbf{29.6}   & \textbf{58.1}   & 78.6            & 104.6           & \textbf{80.3}  & \textbf{97.3}  & \textbf{87.6}   & 85.0           & \textbf{58.7}     & \textbf{112.0}    & 113.9          & 213.9          \\ \hline
                      & \multicolumn{2}{c|}{Purchases}        & \multicolumn{2}{c|}{Sitting}   & \multicolumn{2}{c|}{Sitting Down} & \multicolumn{2}{c|}{Taking Photo} & \multicolumn{2}{c|}{Waiting}    & \multicolumn{2}{c|}{Walking Dog} & \multicolumn{2}{c|}{Walking Together} & \multicolumn{2}{c}{Average}     \\
Milliseconds          & 560                  & 1000           & 560           & 1000           & 560             & 1000            & 560             & 1000            & 560            & 1000           & 560             & 1000           & 560               & 1000              & 560            & 1000           \\ \hline
ConvSeq2Seq\cite{ref14}   & \multicolumn{1}{l}{} &                &               &                &                 &                 &                 &                 &                &                &                 &                &                   &                   &                &                \\
LearnTrajDep\cite{ref12}  & 92.4                 & 125.3          & 78.6          & 109.9          & 88.1            & 137.8           & 78.3            & 95              & 99.0           & 169.5          & 139.2           & 167.7          & 60.3              & 84.1              & 78.9           & 112.6          \\
TrajectoryNet\cite{ref21} & \textbf{84.5}        & 115.5          & \textbf{81.0} & 116.3          & \textbf{79.8}   & \textbf{123.8}  & 73.0            & \textbf{86.6}   & 92.9           & 165.9          & 141.1           & 181.3          & 57.6              & \textbf{77.3}     & 77.7           & 110.6          \\
Ours                  & 86.0                 & \textbf{114.5} & 81.4          & \textbf{116.2} & 82.4            & 126.1           & \textbf{69.8}   & 89.8            & \textbf{91.2}  & \textbf{162.9} & \textbf{135.4}  & \textbf{166.7} & \textbf{55.3}     & 77.6              & \textbf{76.3}  & \textbf{109.6} \\ \hline
\end{tabular}
\end{center}
\end{table*}
\subsection{Results and Discussion}

This section presents the experimental results on the datasets, and a related discuss is provided.

\paragraph{Results on H3.6M.} In Table 1, we compare our model with baselines for short-term prediction. Note that our two-stream model outperforms all the baselines on average prediction errors, demonstrating the effectiveness of our proposed approach. Comparing with the results of convSeq2Seq \cite{ref14} and LearnTrajDep \cite{ref12}, our model predicts much better on movement actions. LearnTrajDep \cite{ref12} respectively model the temporal information and the spatial dependencies of joint trajectories using GCNs, neglecting the co-relation of spatial-temporal features. However, our approach reinforces the spatial-temporal coupling by trajectory spatial-temporal (TST) block, which models joint-level trajectory. Therefore, our model achieves higher accuracy using convolutional TST module along with V-stream dynamic features. Also, comparing with the results of TrajectoryNet \cite{ref21}, our method achieves lower error on some specific common actions such as "Walking", "Eating" and "Smoking". These possible reasons are two folds: (1) The predicted velocity information accumulated to the last frame of the input sequence can keep the motion continuity. (2) Our temporal fusion (TF) module harmoniously keeps the temporal consistency of the P-stream and V-stream feature and better mine the law of the motion due to the fusion of the two streams. So, the accuracy of prediction results combined with two-stream features is higher, especially for regular moves like "Walking" or big move like "Walking Dog".

We provide qualitative comparisons in Fig. 4. Our model effectively captures human motion dynamics and shows better prediction results on "Walking", "Walking Dog". However, the baselines show discontinuity at the first predicting frame, and our method is more natural. These movements are inseparable from the regular swing of the legs so that the velocity vector can retain the displacement of the legs and the velocity stream can model the dynamics of legs more effectively. As for the other motions such as "Direction", "Taking Photo", our model doesn't give a better prediction than TrajectoryNet. To analyze our approach’s failure cases, we visualize the poses in Fig. 5. As for the action "Taking Photo", there are irregular arm movements in these actions that may disturb velocity modeling. Consequently, the irregular arm motion pattern learned from the observation sequence harms the prediction results. However, as shown in "Smoking" for comparison, our model can still give proper prediction in some situ actions.

\begin{figure*}[htbp!]  
	\centering
	\subfigure{
        \includegraphics[width=0.75\textwidth]{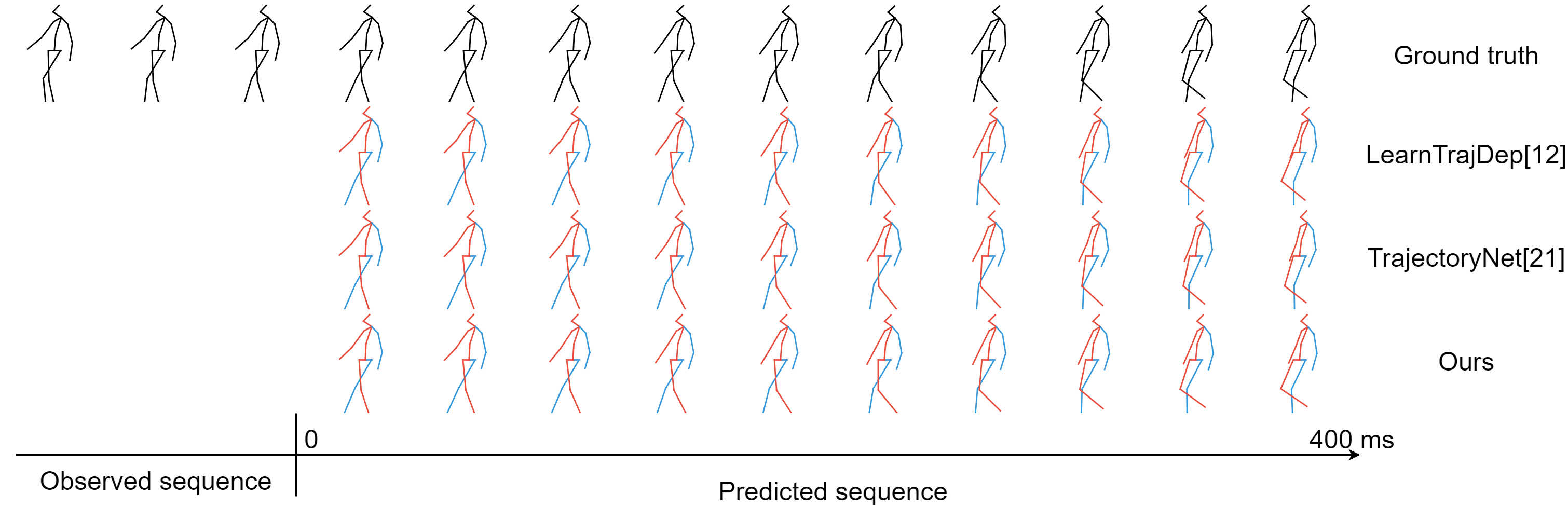}
    }
\end{figure*}
\begin{table*}
(a) Walking
\end{table*}
\begin{figure*}[htbp!]
    \centering
    \subfigure{
        \includegraphics[width=0.75\textwidth]{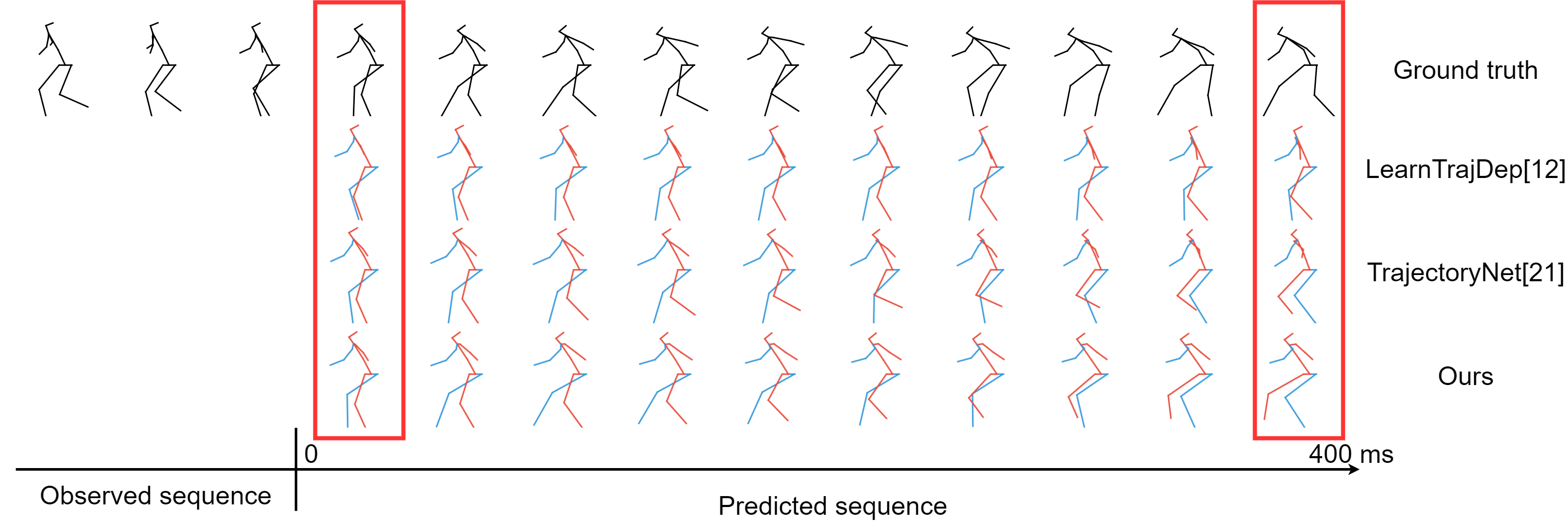}
    }
\end{figure*}
\begin{table*}
(b) Walking Dog
\end{table*}
\begin{table*}
\begin{flushleft}
\textbf{Fig.4:} Quality comparison of "Walking" and "Walking Dog". Both are movement actions. Top: the conditioning sequence and the ground-truth of the predicted sequence. Middle two: state-of-the-art prediction results. Bottom: our prediction. The ground-truth and the input sequences are shown in black. The results evidence that our approach generates high-quality predictions in both cases.
\end{flushleft}
\end{table*}

Our long-term prediction result on H3.6M is shown in Table 2. In long-term experiments, the prediction length is set as 25 frames corresponding to the duration of 1 second. Our model still achieves lower average errors comparing with the baselines. Especially for the movements that perform much better in our model on short-term prediction like "Walking Dog", the long-term prediction still achieves lower error. Therefore, we believe our TF module benefits human dynamics modeling.

\paragraph{Results on 3DPW.} The results of short-term and long-term prediction on 3DPW dataset are shown in table 3. Our method shows better predicting accuracy on short-term predictions. However, the prediction results at 800ms and 1000ms do not exceed the TrajectoryNet \cite{ref21}. 3DPW dataset is a wild action dataset, unlike H3.6M, which has special categories for regular actions such as "Walking" and "Eating", the regularity of 3DPW outdoor actions is complex, so it is difficult for our velocity stream to model the joint movement pattern of the action sequence, and it will accumulate more errors in long-term prediction results.


\begin{figure*}[htbp!]  
	\centering
	\subfigure{
        \includegraphics[width=0.75\textwidth]{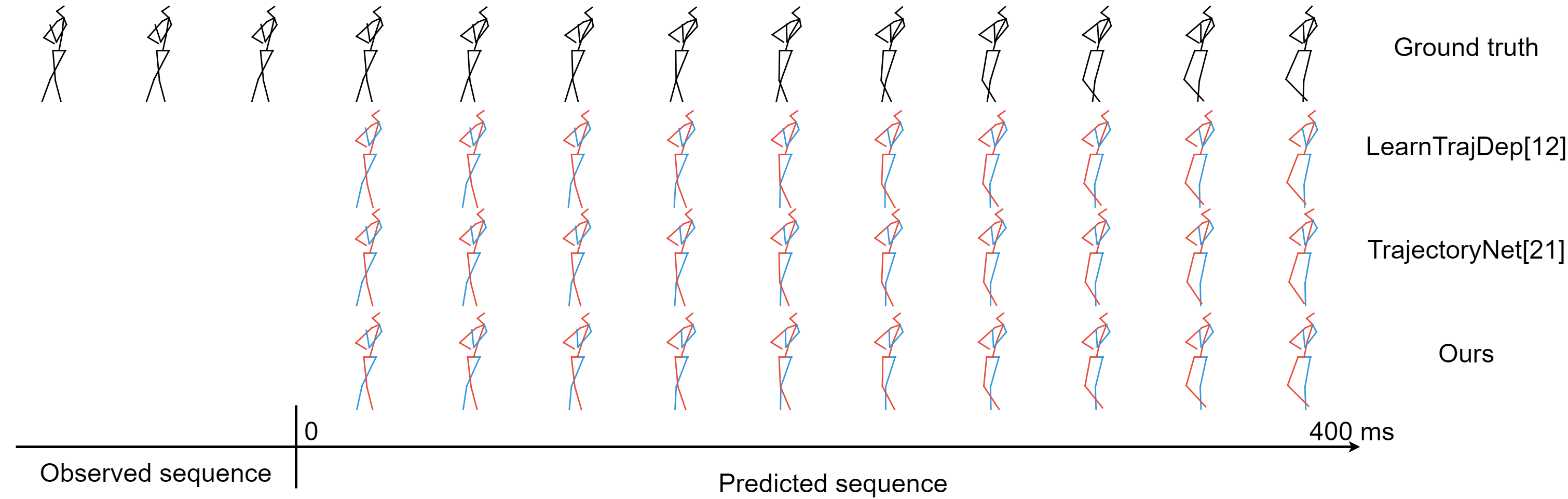}
    }
\end{figure*}
\begin{table*}
(a) Smoking
\end{table*}
\begin{figure*}[htbp!]
    \centering
    \subfigure{
        \includegraphics[width=0.75\textwidth]{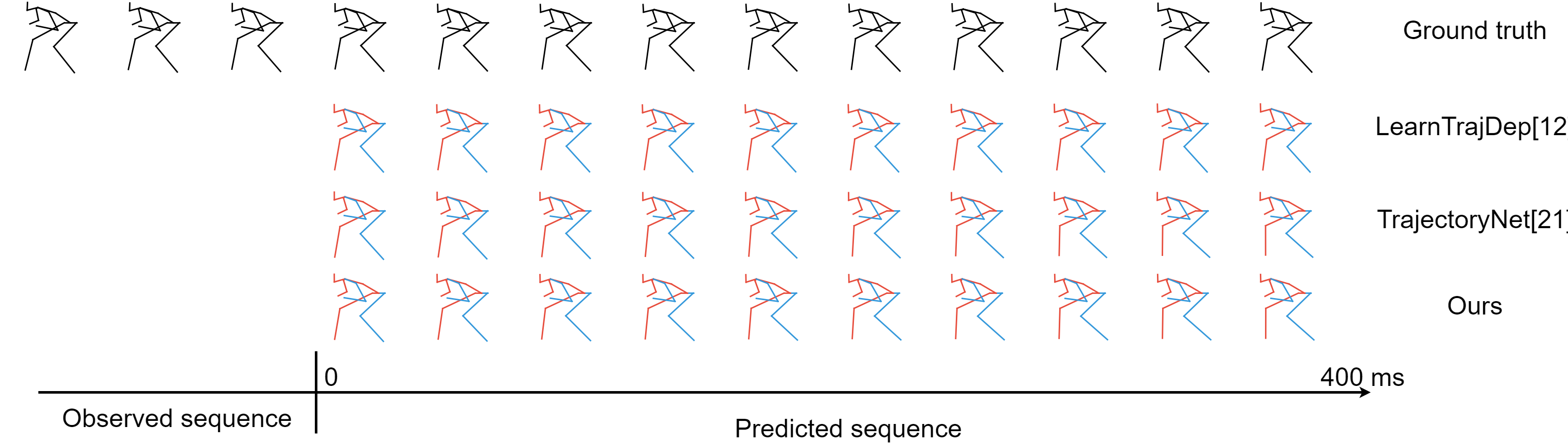}
    }
\end{figure*}
\begin{table*}
(b) Taking Photo
\end{table*}
\begin{table*}
\begin{flushleft}
\textbf{Fig.5:} Quality comparison of "Smoking" and "Taking Photo". Both are in situ actions. Top: the conditioning sequence and the ground-truth of the predicted sequence. Middle two: state-of-the-art prediction results. Bottom: our prediction. The ground-truth and the input sequences are shown in black.
\end{flushleft}
\end{table*}

\begin{table}
\begin{flushleft}
\textbf{Table 3:} Short-term and long-term prediction on 3DPW
\end{flushleft}
\end{table}
\begin{table}[]\normalsize
\begin{center}
\renewcommand\arraystretch{1.3}
\renewcommand\tabcolsep{3pt}
\begin{tabular}{lccccc}
\hline
Milliseconds  & 200           & 400           & 600           & 800           & 1000           \\ \hline
ConvSeq2Seq\cite{ref14}   & 71.6          & 124.9         & 155.4         & 174.7         & 187.5          \\
LearnTrajDep\cite{ref12}  & 35.6          & 67.8          & 90.6          & 106.9         & 117.8          \\
TrajectoryNet\cite{ref21} & 30.0          & 59.7          & 85.3          & \textbf{99.0} & \textbf{107.7} \\
Ours          & \textbf{28.4} & \textbf{59.1} & \textbf{84.6} & 100.0         & 108.2          \\ \hline
\end{tabular}
\end{center}
\end{table}

\paragraph{Results on CMU-Mocap.} Our short-term and long-term prediction result on CMU-Mocap is reported in Table 4. Our method outperforms the baselines on both short-term and long-term predictions. The errors decrease slightly on all time-steps.

\begin{table}
\begin{flushleft}
\textbf{Table 4:} Short-term and long-term prediction on CMU-Mocap
\end{flushleft}
\end{table}
\begin{table}[]\normalsize
\begin{center}
\renewcommand\arraystretch{1.3}
\renewcommand\tabcolsep{3pt}
\begin{tabular}{lccccc}
\hline
Milliseconds  & 80           & 160           & 320           & 400           & 1000           \\ \hline
LearnTrajDep\cite{ref12}   & 11.5          & 20.4         & 37.8         & 46.8         & 96.5          \\
TrajectoryNet\cite{ref21}  & 8.3          & 15.6          & 33.4          & 43.1         & 92.8          \\
Ours & \textbf{8.2}          & \textbf{15.1}          & \textbf{32.8}          & \textbf{43.0} & \textbf{92.6} \\ \hline
\end{tabular}
\end{center}
\end{table}

\paragraph{Discussion.} As shown in Table1~4, our model outperforms the baselines on all time-steps average predictions in H3.6M and CMU-Mocap and underperforms the TrajectoryNet long-term prediction in 3DPW. The reasons are analyzed for two aspects as below.

(1) Dataset analysis. H3.6M and CMU-Mocap record the positional data of human skeleton joints collected by professional equipment indoors. The center of the movement is located at the waist of the human body. Our proposed method can powerfully capture the dynamics of the movement actions and achieve state-of-the-art predictive performance in these datasets. Besides, H3.6M contains more movement actions than CMU-Mocap, which explains our superior results on H3.6M. 3DPW is an outdoor recorded human action dataset. Therefore, the movements of 3DPW are more disturbed and less regular, which makes the prediction more challenging. As mentioned in results on 3DPW, because of the poor movement regularity of 3DPW's long-term action samples, the velocity features interfere with the position space features, so the long-term prediction results do not exceed TrajectoryNet.  

(2) Method analysis. Our proposed two-stream model introduces velocity as an auxiliary input to model human dynamics. The velocity vector can easily capture the regular human motion pattern, which explains our short-term prediction advantage. For further prediction, our model mostly depends on the learned regular pattern which consists of observed velocity features. Therefore, when coupling some irregular movements shot in outdoor locations, the velocity features lead to error accumulating. That explains our superiority in movement actions and disadvantage on some situ actions and irregular actions. After constructing velocity vectors, we use the TF module to fuse the two-stream features chronologically to ensure temporal consistency. In contrast, TrajectoryNet only uses spatial information for modeling, and its dynamic information is not as rich as our two-stream network. Moreover, the multi-scale feature constructed by convseq2seq \cite{ref14} ignores the principle of time consistency.

\begin{table}
\begin{flushleft}
\textbf{Table 5:} Ablation experiments on the different depths of the trajectory spatial-temporal (TST) module. TST-11(Ours) shows the best capability of human motion modeling.
\end{flushleft}
\end{table}
\begin{table}[]\normalsize
\begin{center}
\renewcommand\arraystretch{1.3}
\renewcommand\tabcolsep{3pt}
\begin{tabular}{lcccc}
\hline
Milliseconds & 80           & 160           & 320           & 400           \\ \hline
TST-6        & 10.2         & 23.4          & 49.1          & 59.6          \\
TST-16       & 9.9          & 22.8          & 48.8          & 59.3          \\
TST-21       & 9.9          & 22.8          & 48.7          & 59.2          \\
TST-11(ours) & \textbf{9.8} & \textbf{22.6} & \textbf{48.1} & \textbf{58.4} \\ \hline
\end{tabular}
\end{center}
\end{table}

\subsection{Ablation Analysis}

\paragraph{Evaluation of TST block.} In Table 5, we adopt different convolution depths of spatial-temporal modeling blocks and compare their parameter quantity to confirm our proposed depth's effectiveness. We use residual connections every 5 convolution layers for feature retaining, starting from the first 3*3 convolution layer, noted as a residual block. In the structure of TST, we examine the effect of every residual block. Because of the 3*3 convolution layer at the end of each TST block, the experiment is set to ablation experiment of 6, 11, 16 and 21 convolutional layers. As we can see in Table 5, when using 6 convolution layers (TST-6), the result shows a large margin between the 11 convolution layers (TST-11). In this case, we believe the network doesn't fully capture the global information. However, as the net goes deeper as 16 convolution layers (TST-16) and 21 convolution layers (TST-21), the error doesn't reduce as expected. Therefore, higher parameter quantity and deeper modeling network don't always bring an effect on modeling human dynamics.

As mentioned in Methodology, our TST block's depth is designed by the 22 main joints in H3.6M. The number of joints in 3DPW is 24, which is different from H3.6M. But the joints share nearly the same coordinates at the end of each trunk. Our network can still learn the dynamics of the whole human body. Meanwhile, the number of key joints in CMU-Mocap is 25. When using TST-11, the receptive field of our network doesn't cover all the 25 joints. However, the joints of CMU-Mocap also share nearly the same coordinates at the end of each trunk. Generally speaking, 11 convolution layers can maximize network performance on different datasets.

\begin{table*}
\begin{flushleft}
\textbf{Table 6:} Influence of temporal fusion on H3.6M. Note that, it leads to poorer performance when using concatenation for two-stream fusion instead of temporal fusion.
\end{flushleft}
\end{table*}
\begin{table*}[]\normalsize
\begin{center}
\renewcommand\arraystretch{1}
\renewcommand\tabcolsep{1pt}
\begin{tabular}{l|cccc|cccc|cccc|cccc}
\cline{1-17}
              & \multicolumn{4}{c|}{Walking}                                   & \multicolumn{4}{c|}{Eating}                                     & \multicolumn{4}{c|}{Smoking}                                   & \multicolumn{4}{c}{Discussion}                                 \\
Milliseconds  & 80            & 160           & 320           & 400           & 80            & 160           & 320           & 400            & 80            & 160           & 320           & 400           & 80           & 160           & 320           & 400             \\ \cline{1-17}
TrajectoryNet\cite{ref21} & 8.2           & 14.9          & 30.0            & 35.4          & 8.5           & 18.4          & 37.0            & 44.8           & 6.3           & 12.8          & 23.7          & 27.8          & 7.5          & 20.0            & 41.3          & 47.8            \\
TF:$\times$          & 8.1           & 15.1          & 29.7          & 34.9          & 8.3           & 17.7          & 36.5          & 44.3           & 6.5           & \textbf{12.6} & \textbf{22.3} & \textbf{26.5} & 7.6          & 19.6          & \textbf{39.4} & \textbf{46.5}   \\
TF:$\surd$          & \textbf{7.8}  & \textbf{14.8} & \textbf{28.9} & \textbf{34.4} & \textbf{7.7}  & \textbf{17.1} & \textbf{34.5} & \textbf{42.6}  & \textbf{6.2}  & 12.7          & 23.3          & 27.6          & \textbf{7.1} & \textbf{18.7} & 39.7          & 46.8            \\ \cline{1-17}
              & \multicolumn{4}{c|}{Directions}                                & \multicolumn{4}{c|}{Greeting}                                   & \multicolumn{4}{c|}{Phoning}                                   & \multicolumn{4}{c}{Posing}                                     \\
Milliseconds  & 80            & 160           & 320           & 400           & 80            & 160           & 320           & 400            & 80            & 160           & 320           & 400           & 80           & 160           & 320           & 400             \\ \cline{1-17}
TrajectoryNet\cite{ref21} & 9.7           & 22.3          & \textbf{50.2} & \textbf{61.7} & \textbf{12.6} & 28.1          & 67.3          & 80.1           & 10.7          & 18.8          & 37.0            & \textbf{43.1} & 6.9          & \textbf{21.3} & 62.9          & 78.8            \\
TF:$\times$          & 10.3          & 24.8          & 56.0            & 68.1          & 13.3          & 30.1          & 74.0            & 89.7           & 11.0            & 19.5          & 38.5          & 44.1          & 7.3          & 21.7          & \textbf{62.2} & \textbf{78.0}     \\
TF:$\surd$          & \textbf{9.4}  & \textbf{22.9} & 53.7          & 65.1          & 12.7          & \textbf{28.0}   & \textbf{64.6} & \textbf{79.1}  & \textbf{10.0}   & \textbf{18.6} & \textbf{37.9} & 44.4          & \textbf{6.8} & 21.6          & 63.5          & 79.9            \\ \cline{1-17}
              & \multicolumn{4}{c|}{Purchases}                                 & \multicolumn{4}{c|}{Sitting}                                    & \multicolumn{4}{c|}{SittingDown}                               & \multicolumn{4}{c}{Taking Photo}                               \\
Milliseconds  & 80            & 160           & 320           & 400           & 80            & 160           & 320           & 400            & 80            & 160           & 320           & 400           & 80           & 160           & 320           & 400             \\ \cline{1-17}
TrajectoryNet\cite{ref21} & \textbf{17.1} & \textbf{36.1} & 64.3          & 75.1          & 9.0             & 22.0            & \textbf{49.4} & \textbf{62.6}  & \textbf{10.7} & 28.8          & 55.1          & 62.9          & \textbf{5.4} & \textbf{13.4} & \textbf{36.2} & \textbf{47.0}     \\
TF:$\times$          & 17.8          & 37.2          & 72.4          & 86.7          & 9.4           & 22.4          & \textbf{49.4} & 63.6           & \textbf{10.7} & 28.6          & 56.0            & 67.1          & 5.5          & 13.9          & 36.4          & 47.6            \\
TF:$\surd$          & 17.4          & \textbf{36.1} & \textbf{59.4} & \textbf{67.4} & \textbf{8.6}  & \textbf{21.5} & 50.3          & 63.6           & 11.0            & \textbf{27.4} & \textbf{51.2} & \textbf{60.7} & 5.5          & 13.5          & 37.1          & 49.0              \\ \cline{1-17}
              & \multicolumn{4}{c|}{Waiting}                                   & \multicolumn{4}{c|}{Walking Dog}                                & \multicolumn{4}{c|}{Walking Together}                          & \multicolumn{4}{c}{Average}                                    \\
Milliseconds  & 80            & 160           & 320           & 400           & 80            & 160           & 320           & 400            & 80            & 160           & 320           & 400           & 80           & 160           & 320           & 400             \\ \cline{1-17}
TrajectoryNet\cite{ref21} & \textbf{8.2}  & 21.0            & 53.4          & 68.9          & 23.6          & 52.0            & 98.1          & 116.9          & 8.5           & 18.5          & 33.9          & 43.4          & 10.2         & 23.2          & 49.3          & 59.7            \\
TF:$\times$          & 8.5           & 20.6          & 53.5          & 67.7          & 22.2          & 48.3          & 99.0            & 117.6          & 7.8           & \textbf{17.6} & 34.5          & 42.2          & 10.3         & 23.3          & 50.6          & 61.6            \\
TF:$\surd$          & 8.3           & \textbf{20.5} & \textbf{51.4} & \textbf{66.3} & \textbf{20.8} & \textbf{47.7} & \textbf{94.1} & \textbf{108.7} & \textbf{7.7}  & 18.1          & \textbf{31.9} & \textbf{41.1} & \textbf{9.8} & \textbf{22.6} & \textbf{48.1} & \textbf{58.4}   \\ \cline{1-17}
\end{tabular}
\end{center}
\end{table*}

\paragraph{Evaluation of TF module.} We use direct concatenation instead of temporal fusion for fusing two-stream features from position space and velocity space. When implementing velocity vector without temporal fusion(denoted as TF:$\times$ in Table 6), the model shows a higher error on most prediction results. Comparing with TrajectoryNet, which models position space dynamics only, there's still a gap on average prediction on all time steps. In this case, we believe the features from velocity space disturb the position space modeling when using concatenation. Because of the lack of temporal consistency of the two-stream features through direct concatenation, the network doesn't model the position and velocity information of the human body at the same time-step when forward modeling, but treats it as a single posture evolution along the time sequence leading to spatial-temporal in conformity. Therefore, abandoning temporal fusion only makes the velocity vector interfere with the prediction result of position space.

\section{Conclusion}

In this paper, we propose a TF module based two-stream architecture that models position stream and velocity stream features for human motion prediction. To ensure the integration of spatial-temporal co-occurrence, the TF module fuses the features from two streams in chronological order, which maintains temporal consistency and shows an effect on feature fusion comparing with related works. Meanwhile, the introduced high-dimensional information, which is the velocity vector, shows its advantage on both short-term modeling and long-term modeling for movement action predicting. Our future work will focus on the TF module generalization that can be adapted to the fusion phase of other two-stream or multi-stream modeling networks.

\end{sloppypar}
\pagestyle{empty} 
\end{document}